%% file: main.tex
\documentclass[runningheads]{llncs}

\usepackage{eccv}

\usepackage{eccvabbrv}

\usepackage{graphicx}
\usepackage{booktabs}

\usepackage{amsmath}
\usepackage{parskip}
\usepackage{pgffor}
\usepackage{subcaption}
\usepackage{xurl}

\usepackage[accsupp]{axessibility}  %

\usepackage{hyperref}

\usepackage{orcidlink}

\input{preamble}

\begin{document}

\title{Training-Free Reconstruction-Based AI-Generated Image Detectors Are Inherently Vulnerable to Adversarial Examples}

\titlerunning{Training-Free Rec.-Based AIGI Detectors Are Inherently Vuln.\ to Adv.\ Ex.}

\author{Roman Demchenko\,\orcidlink{0009-0001-8105-5219} \and
Jonas Ricker\,\orcidlink{0000-0002-7186-3634} \and
Asja Fischer\,\orcidlink{0000-0002-1916-7033}}

\authorrunning{R. Demchenko, J. Ricker, A. Fischer}

\institute{Ruhr University Bochum, Bochum, Germany \\
\email{\{roman.demchenko,\,jonas.ricker,\,asja.fischer\}@rub.de}}

\maketitle

\begin{abstract}
The impressive visual quality and ubiquity of AI-generated images call for reliable and robust detection methods.
Reconstruction-based detectors have emerged as a promising direction for transparent and training-free identification of synthetic images.
However, due to their fundamentally different mode of operation (compared to standard, classifier-based methods), little is known about their adversarial robustness.
In this work, we propose two novel attack methods targeted at detectors that leverage autoencoder reconstruction error.
We find that by constructing imperceptible adversarial examples, the distance between original and reconstruction can be artificially increased, causing fake images to be wrongly classified as real.
Our evaluation including images from three state-of-the-art generators and three detectors demonstrates that detection performance is significantly decreased, even if attacked images additionally undergo real-world degradations.
Critically, our adversarial examples naturally transfer across detectors, as they all share the same principle, pointing towards an inherent vulnerability of reconstruction-based detectors.

\end{abstract}

\section{Introduction}
\label{sec:intro}

Image generation models have advanced significantly in recent years. Modern latent diffusion models (LDMs)~\cite{ldm} are capable of producing high-quality synthetic images. The spread of online services such as Midjourney \cite{midjourney} and DALL-E \cite{dalle} allows for simple and fast production of synthetic images without any technical knowledge. Additionally, open-source models like Stable Diffusion \cite{stable-diffusion} and Flux \cite{flux} can be run on local machines with consumer-grade hardware, without limitations by third-party services or regulations.

The large number of easily accessible and highly capable image-generating models raises concerns about the associated risks and threats they may bring. For example, synthetic images can be used for creating fake identities on social media platforms \cite{fakeprofiles}, facilitating fraud \cite{turkey}, or spreading fake news \cite{pope}. This creates an urgent need for reliable detection methods.

Reconstruction-based detectors~\cite{aeroblade,hfi,rdd} have recently gained attention as a promising approach for detecting LDM-generated images. Unlike the majority of existing detectors, they do not train a neural network-based classifier on real and generated images. Instead, they are based on the assumption that the autoencoder (AE) used by a specific LDM can reconstruct generated images more accurately than real images. Consequently, images with low reconstruction error are classified as generated (by the LDM corresponding to the AE), while images with higher error are classified as real.
By combining different AEs in an ensemble and using the minimum error for each respective image, detectors can generalize to a range of different LDMs.
This detection approach provides several advantages. It requires no costly training, can be easily adapted to novel generative models (if the AE is available), and is more interpretable due to not relying on a black-box classifier.

Outside controlled settings, however, detection performance may be affected by both natural and adversarial image modifications.
While most detectors are commonly evaluated for robustness against common image degradations, robustness against adversarial examples has not yet been extensively studied.
Prior work shows that conventional data-driven detectors are highly susceptible to both white- and black-box attacks~\cite{adversarialrobustness,adversarialrobustness2}.
Given that reconstruction-based approaches lack a dedicated classifier, the question arises whether such detectors offer increased adversarial robustness.

In this work, we provide the first comprehensive adversarial robustness analysis of reconstruction-based AI-generated image detectors. Unlike methods incorporating trained classifiers, computing adversarial examples against training-free detection methods is not straightforward due to the absence of the target model. 
We therefore develop two novel attacks. Our first attack operates in latent space, causing an image's embedding to deviate from the original. Secondly, we directly target a differentiable distance function in pixel space.
It should be noted that both attacks do not require detailed knowledge of the detection methods, but instead directly target the underlying idea of fake images yielding lower reconstruction errors.
Our empirical evaluation including three detectors and three datasets demonstrates that both attacks consistently degrade detection performance with minimal impact on visual quality.
We also observe that attacks are transferable across AEs and remain effective when images undergo typical degradations like compression and blurring.

In summary, we make the following contributions:
\begin{itemize}
    \item We discover that training-free, reconstruction-based AI-generated image detectors are inherently vulnerable to adversarial examples. This vulnerability is not limited to specific detectors, but universally applies to the principle of training-free reconstruction-based detection.
    \item We explore two novel attack vectors, demonstrating their effectiveness across three detectors while remaining imperceptible to human observers.
    \item Our evaluation confirms that both attacks are robust to the choice of AEs and common image degradations, making them a considerable threat in realistic settings.
\end{itemize}

We release our code and data at \url{https://github.com/romandemchenkox/trainingfree-reconstruction-based-detectors-are-vulnerable-to-adversarial-examples}.

\section{Related Work}

\subsection{AI-Generated Image Detection}
Since generative models are able to create realistic-looking images, distinguishing real from generated images is an active area of research.
Different detection approaches have emerged over time.
While early methods exploit high-level artifacts or semantic inconsistencies~\cite{maternExploitingVisualArtifacts2019,huExposingGANGeneratedFaces2021,guoEyesTellAll2022}, other approaches leverage low-level generation artifacts in the pixel~\cite{mccloskeyDetectingGANgeneratedImagery2019,barniCNNDetectionGANGenerated2020} or frequency domain~\cite{zhangDetectingSimulatingArtifacts2019,frankLeveragingFrequencyAnalysis2020}.
The dominant detection approach, however, is to train a neural network-based classifier on real and generated images, thereby automatically learning relevant features~\cite{wangCNNgeneratedImagesAre2020,mandelliDetectingGANgeneratedImages2022}.
It has recently been shown that using pretrained vision foundation models such as CLIP~\cite{radfordLearningTransferableVisual2021} as feature extractors can improve performance and generalization to unseen generative models~\cite{ojhaUniversalFakeImage2023,koutlisLeveragingRepresentationsIntermediate2025}

Reconstruction-based methods~\cite{aeroblade,hfi,rdd} work fundamentally different, in particular due to not requiring any training.
They are based on the intuition that a generative model ``remembers'' the images it generated, allowing it to reconstruct them more accurately than real images.
In its pure form, classification is performed by computing the distance between the original and reconstruction, with low errors suggesting that an image is AI-generated.
AEROBLADE~\cite{aeroblade} applies this idea to LDMs by reconstructing images solely using the model's AE.
HFI~\cite{hfi} and RDD~\cite{rdd} further improve detection performance and generalization by reducing biases regarding image background and additionally leveraging latent space reconstruction errors, respectively.
Another class of reconstruction-based methods, like FakeInversion~\cite{fakeinversion} and DIRE~\cite{dire}, leverages the entire diffusion and denoising processes for reconstruction.
The differences between the original image and its reconstruction are then used as training data for a neural network-based classifier.

We limit our evaluation to methods that directly use reconstruction errors for detection, without relying on a trained classifier.

\subsection{Adversarial Robustness of AI-Generated Image Detectors}
Compared to the number of works proposing novel detection methods, relatively little research is conducted on their adversarial robustness.
The pioneering work by Carlini and Farid~\cite{adversarialrobustness} demonstrates that neural network-based classifiers become ineffective when adding imperceptible perturbations.
While attacks are most effective in a white-box setting, 
the AUROC can be reduced from 0.95 to 0.22 in a black-box setting using a surrogate classifier.
The general vulnerability is confirmed by subsequent works and extended to more recent detectors~\cite{adversarialrobustness2,derosaExploringAdversarialRobustness2024}.
However, for training-free reconstruction-based detectors, a detailed and targeted robustness analysis is still lacking. %
While AEROBLADE~\cite{aeroblade} was previously evaluated under an attack derived from a surrogate CNN model~\cite{aerobladeadv}, it remains unclear whether the core rationale of reconstruction-based detection---the reconstruction error---can be attacked.

\section{Preliminaries}

In the following, we briefly recap the architecture of LDMs, introduce the three reconstruction-based detection methods we evaluate, and explain the concept of the chosen gradient-based method for computing adversarial examples. %

\subsection{Latent Diffusion Models}
Latent diffusion models (LDMs)~\cite{ldm} are generative models capable of producing synthetic images conditioned on text prompts and, in some cases, additional inputs, such as images.
LDMs typically consist of two main components: the U-Net \cite{unet} modeling the diffusion process and an autoencoder (AE). In contrast to DMs, which operate in pixel space, an image processed by an LDM is first encoded into a latent representation by the AE. The U-Net is then trained on latents of real images, in which Gaussian noise is gradually applied to the input latent during the forward diffusion process, allowing the model to learn to remove the added noise and restore the original latent. After training, the model is capable of producing fully synthetic images from Gaussian noise, guided by a text prompt. This process is called denoising. After denoising, the resulting latent is decoded back to the pixel space, forming the final image.

\subsection{Detection Methods}

Different reconstruction-based detection methods for LDM-generated images have been proposed \cite{aeroblade,hfi,rdd}.
All of them have in common that they allow identifying generated images stemming from a certain model based on an analysis of their AE reconstruction.
In the following, we briefly introduce three state-of-the-art, training-free approaches in more detail.

\textbf{AEROBLADE}~\cite{aeroblade} reconstructs the input image using the AE of an LDM and compares it to the original. The LPIPS \cite{lpips} metric, which measures perceptual similarity of two images, is used to capture the differences between the image and its reconstruction. The LPIPS distance between them is observed to be higher for real images than for fake images generated by the same LDM. This is because synthetic images from the LDM are generated from latents that lie within the AE's learned latent distribution. As a result, they can be reconstructed accurately by the same AE. However, images that were not generated by the same LDM are not guaranteed to be well represented by this latent distribution. Consequently, reconstructing them may introduce distortions, which are highlighted by higher LPIPS values. The reconstruction error computed by AEROBLADE can be formulated as follows:
\begin{equation}
\Delta_\text{AEROBLADE} = d(x, \aeix{i}{x}) \enspace,
\end{equation}
where  $\aeix{i}{\cdot} = \mathcal{D}_i(\mathcal{E}_i(\cdot))$ denotes the AE consisting of an encoder $\mathcal{E}_i$ and a decoder $\mathcal{D}_i$, %
and $d$ is a distance metric, which is commonly chosen to be LPIPS.

\textbf{HFI}~\cite{hfi} operates similarly but computes the reconstruction error in a more robust way. 
The authors observe that high-frequency components of an image have the greatest influence on its reconstruction error. To isolate the contribution of high-frequency components to the final prediction, the reconstruction error is computed separately for the original image and a version where high-frequency components are removed. This is done with a low-pass filter function, such as Gaussian blur. The LPIPS distance between the low-pass filtered image and its reconstruction is subtracted from the distance of the original. The reconstruction error of HFI is defined as:
\begin{equation}
\Delta_\text{HFI} = d(x, \aeix{i}{x}) -  d(\mathcal{F}(x), \aeix{i}{\mathcal{F}(x)})
\end{equation}

\textbf{RDD}~\cite{rdd} is the most recent detector and an extension of HFI. It combines debiased reconstruction errors in both pixel and latent space. The reconstruction error in the pixel space is measured by the $S_\text{image}$ score. This score is obtained by combining two debiasing strategies, based on the ideas applied in HFI. First, the reconstruction error of an image is compared to that of its rotated version. The intuition is that rotating real images does not significantly change their reconstruction error, while rotating fake images makes them more difficult to reconstruct. Second, a low-pass filter is applied to suppress the contribution of simple backgrounds, isolating only high-frequency components of the images. The reconstruction error in the latent space is represented by the $S_\text{latent}$ score. It measures the accuracy of the reverse diffusion process of the U-Net of the chosen LDM on the input latent and compares it to that of a rotated latent. The intuition behind this approach is that rotating latents of images that are unknown to the diffusion model cannot be denoised as accurately as latents of real images or images generated by that particular LDM.

The final score of RDD is the combination of $S_\text{image}$ and $S_\text{latent}$:
\begin{equation}
\Delta_\text{RDD} = S_\text{image}(x) \times {S_\text{latent}(\enci{i}{x})}^2 \enspace .
\end{equation}
All three detection methods can be configured with multiple AEs by taking the minimum reconstruction error across all AEs as the final detection score.

\subsection{Adversarial Examples}
Suppose a neural network $f_\theta$, given an input $x$, produces a prediction $\hat{y} = f_\theta(x)$, which matches the ground truth label $y$. The core idea of an adversarial attack is to find a sample $\adv{x}$ that is reasonably similar to $x$, so that the target model misclassifies it. This condition can be defined as follows:
\begin{equation}
f_\theta(\adv{x}) \neq y, \quad \text{ s.t. } \enspace d(x, \adv{x}) \leq \epsilon \enspace .
\end{equation}
In particular, this can be achieved by means of gradient-based attacks, such as projected gradient descent (PGD)~\cite{pgd}. PGD generally targets a neural network and computes gradients of the chosen loss with respect to the model input. Then it applies perturbations to the input, so that the chosen loss is maximized and the changes remain relatively imperceptible. The strength of the perturbation is controlled by the chosen perturbation budget $\epsilon$.

\section{Threat Model}
\label{sec:threat}

Due to the distinct nature of training-free reconstruction-based detection methods, our threat model differs from the ``classical'' adversarial setting.
Typically, the attacker's goal is to create an adversarial example that causes a classifier to wrongly predict a label that differs from the ground-truth label.
In a white-box setting, the attacker has access to the architecture and weights of the classifier.
In contrast, we refer to a black-box setting if the attacker can only query the classifier or uses a surrogate model to craft adversarial perturbations.

However, training-free reconstruction-based AI-generated image detectors do not utilize a dedicated classifier.
Instead, the autoencoder of a generative model (or multiple ones) essentially \emph{is} the classifier.
Classical definitions of white- and black-box attacks are therefore not suitable, since the attacker has inherent knowledge about the defender's abilities (except for the specific set of AEs in the ensemble).
In the defender's best-case scenario, the AE that generated the image is part of the set of AEs used for detection, as it should yield the lowest reconstruction error.
Since the attacker knows which AE generated the image (and has access to it), they naturally have white-box access and can craft an ideal adversarial perturbation.
In case the defender's set of AEs does not include the AE that created the attacker's image, there are two outcomes in which the attacker succeeds. 
Either the image is not classified as fake (even without adversarial perturbations), since none of the candidate AEs yields a low enough reconstruction error.
Alternatively, if the ``next-best'' AE in the ensemble can produce an accurate reconstruction, the adversarial example may generalize (as we show in Section~\ref{sec:transferability}) and still cause a misclassification.

\section{Attacks on Reconstruction-Based Detectors}

Gradient-based attacks, like PGD, generally require a differentiable model to be able to compute a gradient with respect to the given input. Training-free detection methods do not incorporate a trained classifier dedicated to distinguishing between fake and real images. As a consequence, gradient-based attacks cannot be applied in the same manner as for conventional classifier-based methods. However, their reconstruction process still involves differentiable components. Training-free reconstruction-based detectors rely on the assumption that images produced by a particular LDM can be reconstructed more easily using the matching AE. Since an AE is differentiable, it is a suitable target for a gradient-based attack.

The core idea of the proposed attack method is to disrupt the reconstruction process, making detection methods unable to distinguish between perturbed fake images and real images. This can be accomplished by adding an adversarial perturbation to the input image, causing a higher reconstruction error.

\subsection{Latent Space Attack}
One way to achieve this goal is to modify the target image in a way that maximizes the distance between the latent representations of the attacked and original image within the latent space of the respective AE. %
This corresponds to maximizing the following loss function w.r.t.~$\eta$,
\begin{equation}
L_\text{latent}(x, \eta) = \lVert \mathcal{E}(x) - \mathcal{E}(x+\eta) \rVert_2  \enspace,
\end{equation}
where $\mathcal{E}(\cdot)$ represents the encoder of the given AE, $\lVert \cdot \rVert_2$ the Euclidean norm, and $\eta$ is from a predefined $\epsilon$-region to ensure the adversarial image is reasonably similar to the original image. %
Intuitively, the largest change in latent space can be achieved by leaving the data manifold, which makes the image harder to reconstruct by the decoder, leading to an increase in reconstruction error.

\subsection{Pixel Space Attack}

An alternative approach is to increase the distance between the image and its reconstruction directly. When measuring the distance in terms of, for instance, LPIPS, this is equivalent to maximizing 
$
L_\text{pixel}(x, \eta), \text{ s.t. } \lVert\eta  \rVert_\infty < \epsilon ,
$
with
\begin{equation}
L_\text{pixel}(x, \eta) = \text{LPIPS}(x+\eta, \mathcal{D} \circ \mathcal{E} (x+\eta)) \enspace,
\end{equation}
where $\mathcal{E}(\cdot)$ and $\mathcal{D}(\cdot)$ denote the encoder and decoder of the AE, respectively. Since both LPIPS and the AE are differentiable, the gradients of the loss function can be computed via backpropagation. Note that LPIPS could be replaced by any other choice of distance measure, as long as it is differentiable.

\section{Experiments}

We first describe how we conducted our experiments, after which we evaluate our proposed attacks. Lastly, we analyze the quality of our adversarial examples and perform different ablations.

\subsection{Experimental Setup}

We evaluate our attacks on a subset of the Synthbuster dataset \cite{synthbuster}, which contains synthetic images generated by several LDMs. The real images are taken from the RAISE-1k dataset \cite{raise}. This dataset contains high-quality photos of different scenes, based on which all images from the Synthbuster dataset were generated to ensure semantic alignment.
Two additional sets of images were generated using Flux.1-schnell and Stable Diffusion (SD) 3.5-medium, based on the prompts provided in the Synthbuster dataset and using default parameters for each LDM.
To match the employed AEs, we only include images generated by Midjourney-v5, Flux.1-schnell, and SD3.5 in the experiments. According to prior work \cite{hfi}, using the SD2 AE tends to produce more accurate predictions in reconstruction-based detection methods on Midjourney than on the actual SD2 dataset. In total, we use 300 synthetic and 100 real images.
All images are preprocessed to a fixed resolution of $512\times512$ pixels using center cropping and resizing. This provides a reasonable compromise between image quality, expected input resolution of the employed models, and computational cost.

 The attack targets reconstruction-based training-free detection methods. Therefore, three state-of-the-art detectors are selected for evaluation. AEROBLADE and HFI are initialized with the distance function $\text{LPIPS}_2$ (referring to its second layer), as it has been shown to be the most effective for both methods~\cite{hfi}. For HFI, a Gaussian blur with a $3 \times 3$ kernel and a standard deviation of $\sigma=0.8$ is used as the low-pass filter. RDD applies the same Gaussian blur settings and a 90° rotation for the respective transformations. All parameters are set in accordance with the original publication, namely $\lambda_R=0.5$ and $\lambda_L=1$. For better comparability between detectors, $\text{LPIPS}_2$ is also used as the distance function. Since only LDM-generated images are used in the experiments, the $S_{\text{image}}$-score of RDD is used as its final prediction.
 All detectors are evaluated using an ensemble of AEs corresponding to SD2, SD3.5, and Flux. Following prior work, the minimum reconstruction error across the selected AEs is used as the final prediction.
 
Unlike trained classifiers, training-free methods do not automatically provide a decision threshold. Their score distributions vary across different datasets and often overlap between real and fake images, making threshold-dependent metrics unreliable. The performance is therefore measured using the area under the ROC curve (AUROC), which evaluates the ability of a detector to distinguish between real and fake images across all decision thresholds.
All variants of the proposed attack are run using a fixed number of iterations $n= 20$ and a step size $\alpha=\frac{2\epsilon}{n}$, where $\epsilon$ is the $L_{\infty}$ perturbation budget. The attack is evaluated with $\epsilon$ ranging from $1/255$ to $10/255$. Furthermore, a random start within the $L_\infty$-ball of radius $\epsilon$ around the original image is employed.

\subsection{Evaluating Attack Strength}
\begin{figure}[t]
\centering
\begin{subfigure}{0.49\columnwidth}
\centering
\includegraphics[width=\linewidth]{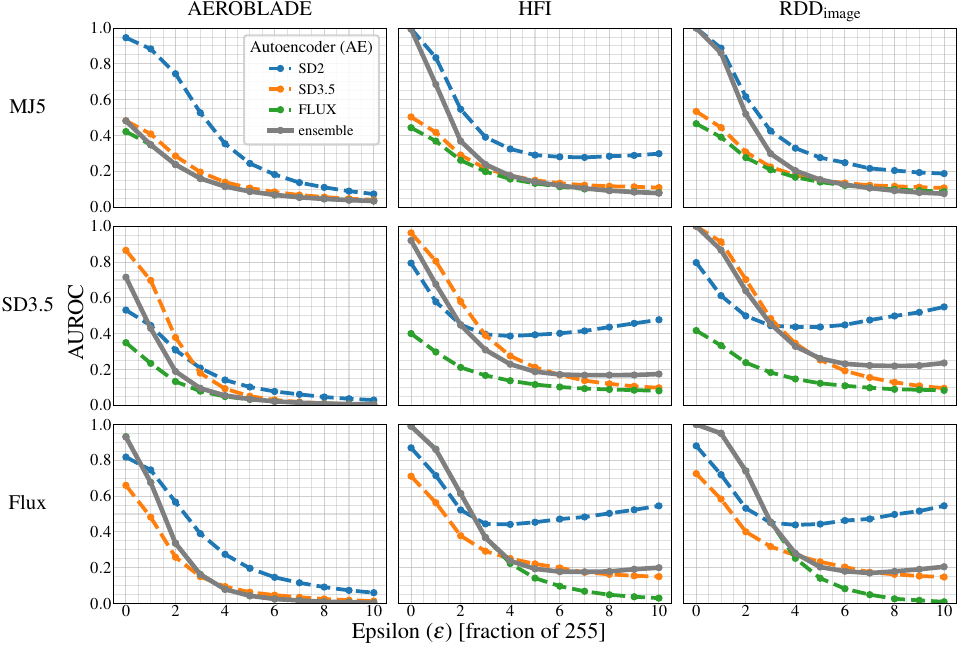}
\caption{\lossA}
\end{subfigure}
\hfill
\begin{subfigure}{0.49\columnwidth}
\centering
\includegraphics[width=\linewidth]{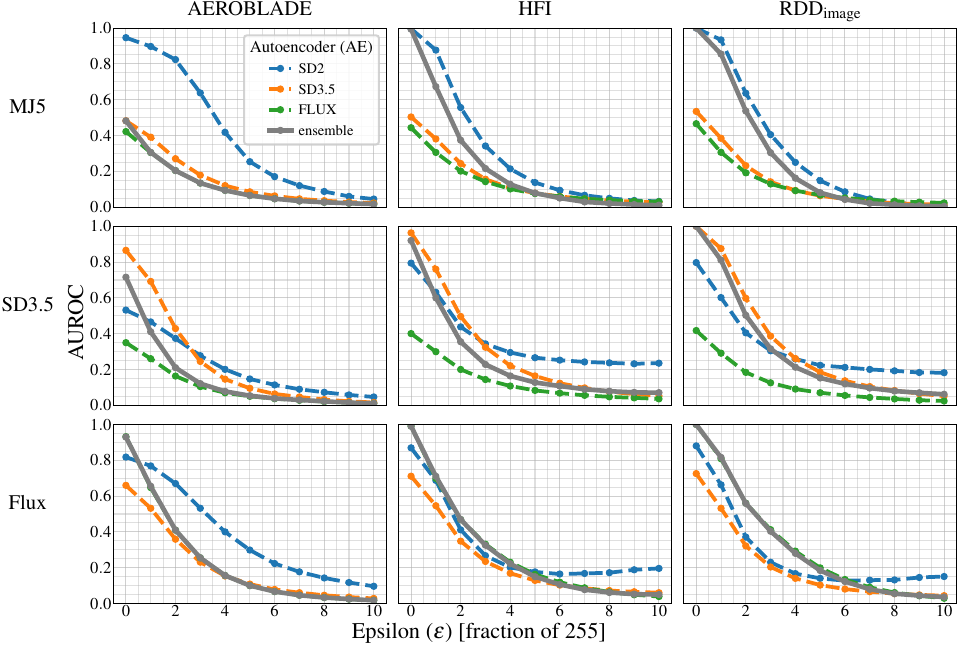}
\caption{\lossB}
\end{subfigure}
\caption{AUROC of the detectors under the proposed attack on the three selected datasets.}
\label{fig:pgd-auroc}
\end{figure}

The detection performance is first measured on clean images (i.e., $\epsilon = 0$) and then under the proposed attack at different $\epsilon$-budgets. To mimic the practically relevant setting, only fake images are perturbed. The attack is computed independently of the target detector, meaning the same images are evaluated by all detectors. The AUROC under the attack targeting the matching AEs on the three selected datasets is presented in \fig{pgd-auroc}. The impact of each attack variant (denoted as \lossA and \lossB) is shown on a grid of nine plots, where each plot corresponds to a detector-dataset pair.

As expected, an AE performs better on the matching dataset than on non-matching datasets. For example, the SD2 AE detects fake images from Midjourney with high confidence, whereas the Flux AE is only reliable for images generated by Flux. The plots show that this trend is consistent across all AEs and detectors. Notably, the ensemble performance is not always as high as the performance of the best AE. This effect could be attributed to Flux and SD3.5 generally producing lower reconstruction errors than older AEs \cite{flux}, making an ensemble approach more fragile.

The AUROC of all three detectors decreases substantially with increasing $\epsilon$ for all AEs on all datasets, indicating the effectiveness of the attack. The performance of AEROBLADE decreases monotonically for both attack variants. HFI behaves similarly under \lossB, but its performance with the SD2 AE stabilizes around 0.4 after $\epsilon=4/255$ under \lossA, slightly drifting upwards as $\epsilon$ increases. A similar pattern is observed for RDD as well under \lossA. We observe that \lossB is significantly more effective than \lossA. Additionally, for each dataset, an attack based on the matching AE is most effective against the respective AE used for detection. Generally, detection becomes unreliable (AUROC $\le0.5$) at $\epsilon=3/255$ or $\epsilon=4/255$ for the three detectors using any of the AEs for estimating the adversarial example on all datasets. Among the considered detection methods, RDD appears slightly more robust, retaining the highest ensemble performance.

Based on our results, we conclude that the attack is highly effective against the selected detectors, with HFI being more robust than AEROBLADE, and RDD demonstrating the highest robustness among them.

\subsection{Transferability}
\label{sec:transferability}
To better understand the real-world threat of the attack (i.e., the threat in a scenario in which the attacker assumes a reconstruction-based detection method, but does not know which AE is used),
it is important to assess the transferability of the attack between different AEs. For that, the attack is run using the AEs of SD2, SD3.5, and Flux and evaluated against the detectors initialized with all three of them. Performance is averaged across the selected datasets.

\begin{figure}[t]
\centering
\begin{subfigure}{0.49\columnwidth}
\centering
\includegraphics[width=\linewidth]{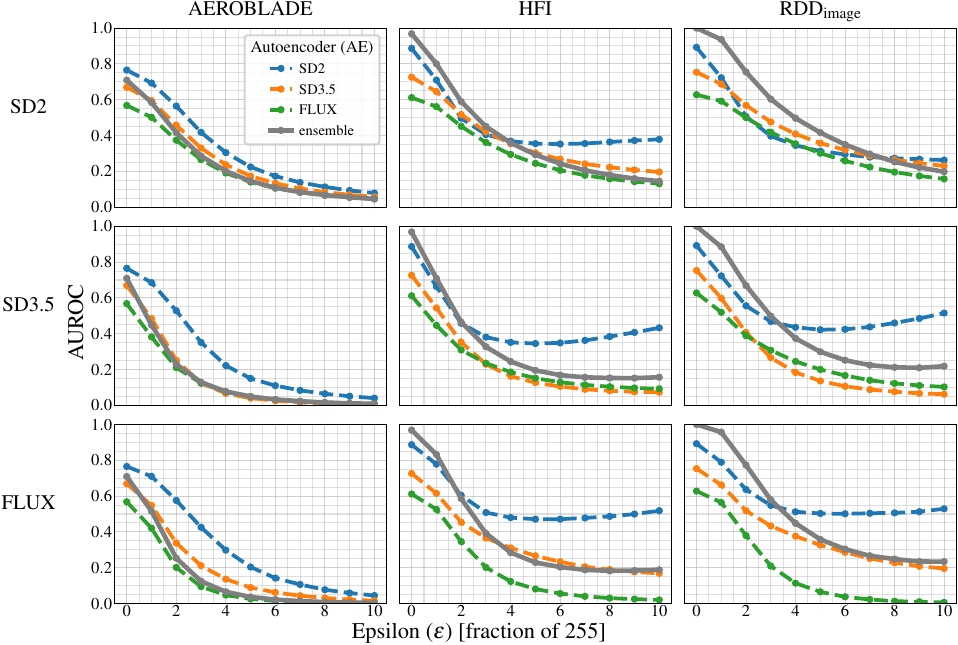}
\caption{\lossA}
\end{subfigure}
\hfill
\begin{subfigure}{0.49\columnwidth}
\centering
\includegraphics[width=\linewidth]{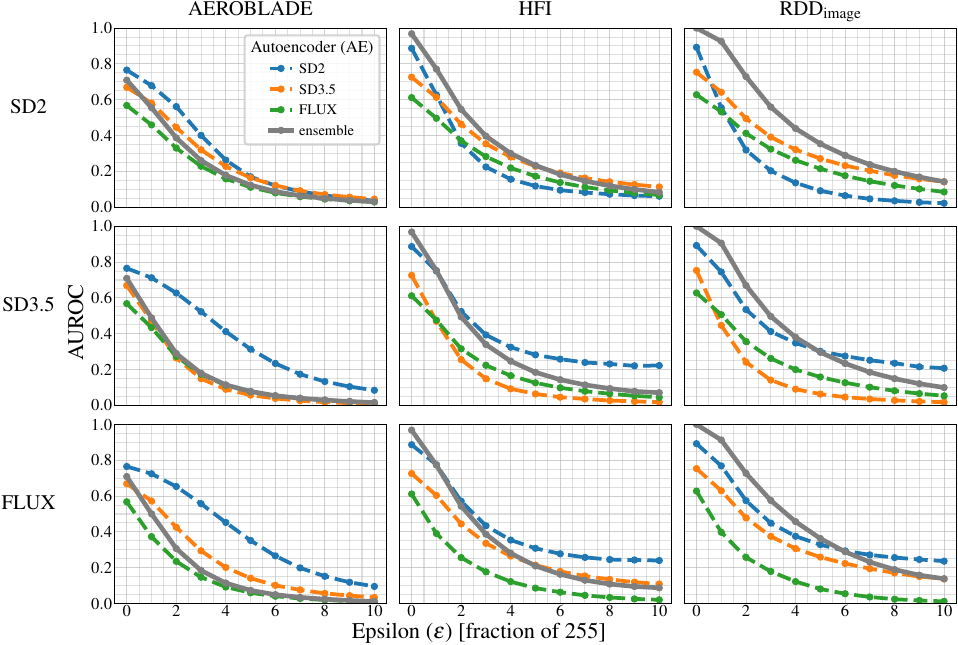}
\caption{\lossB}
\end{subfigure}

\caption{AUROC of the detectors under the proposed attack targeting the selected AEs.}
\label{fig:transfer1}
\end{figure}

As observable in the results shown in \fig{transfer1}, the attack remains effective against all three detectors, even when targeting only a single AE. %
Interestingly, detectors using SD2 only tend to have the highest performance compared to other AEs, no matter which AE is targeted by the attacker. When SD3.5 or Flux are targeted, the respective AE consistently demonstrates the lowest performance.

Overall, the attack using \lossB is more effective than \lossA in this setting. Furthermore, the attack exhibits strong transferability across different AEs.

\subsection{Quality of Adversarial Examples}

\begin{figure}[t]
\centering
\begin{subfigure}{0.49\columnwidth}
\centering
\includegraphics[width=\linewidth]{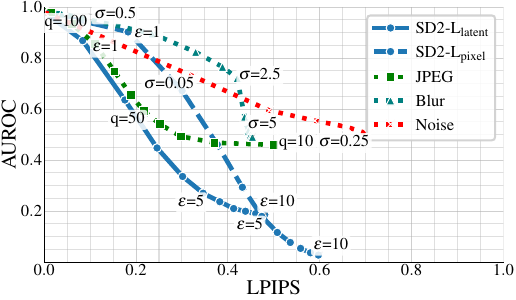}
\caption{LPIPS}
\end{subfigure}
\hfill
\begin{subfigure}{0.49\columnwidth}
\centering
\includegraphics[width=\linewidth]{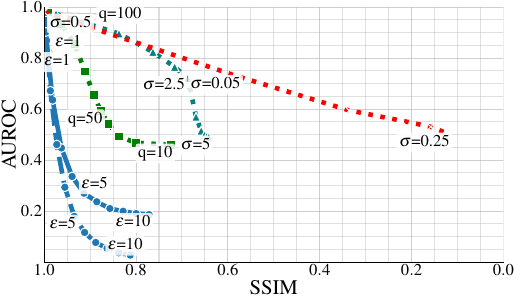}
\caption{SSIM}
\end{subfigure}
\caption{AUROC of different attack variants and distortions in relation to the image quality.}
\label{fig:lpips}
\end{figure}

Clearly, larger perturbations lead to diminishing detector performance. However, too large perturbations may attract attention and expose the fact that the image has been tampered with. Therefore, it is important to consider the imperceptibility of the computed perturbations.
To assess it, the adversarial perturbations are compared to regular transformations. Imperceptibility is evaluated using LPIPS and structural similarity index measure (SSIM)~\cite{wangImageQualityAssessment2004a} by measuring the similarity between the original and perturbed images. \fig{lpips} shows the mean distance values of both variants of the attack targeting the SD2 AE and tested on the Midjourney dataset.

The attack generally yields a better AUROC-LPIPS tradeoff than natural corruptions such as JPEG, Gaussian blur, and noise. JPEG demonstrates lower LPIPS values despite introducing significant artifacts. \lossA drops below AUROC=0.5 for $\epsilon=3$ while staying at an LPIPS value of about 0.24. \lossB shows the same AUROC but a far larger LPIPS value of about 0.38 at the same value of $\epsilon$. SSIM, on the other hand, is relatively high for both attack variants, staying around 0.96 at $\epsilon=3$.

Overall, it can be said that the proposed attack remains reasonably imperceptible with \lossA and becomes more apparent for \lossB.

\subsection{Adversarial Examples in Real-World Settings}

While adversarially forged images are effective when evaluated by the target detection methods directly, this may not be the case in a practical setting. Images are compressed, resized, or otherwise distorted when uploaded to the Internet. Studying how such transformations may affect the impact of the adversarial perturbations is essential to assessing their actual threat in practice.

\begin{figure}
\centering
\begin{subfigure}{0.49\columnwidth}
\centering
\includegraphics[width=\linewidth]{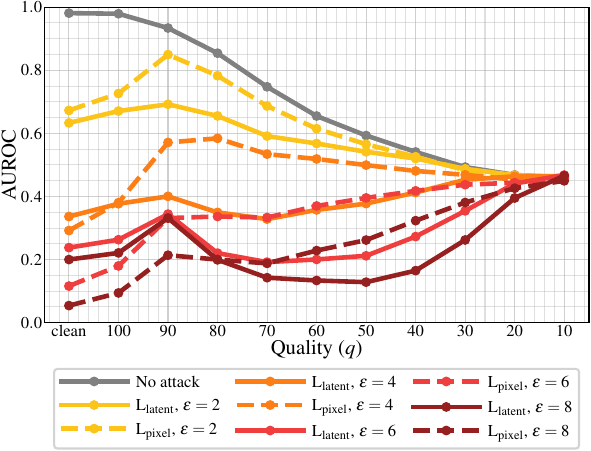}
\caption{JPEG}
\end{subfigure}
\hfill
\begin{subfigure}{0.49\columnwidth}
\centering
\includegraphics[width=\linewidth]{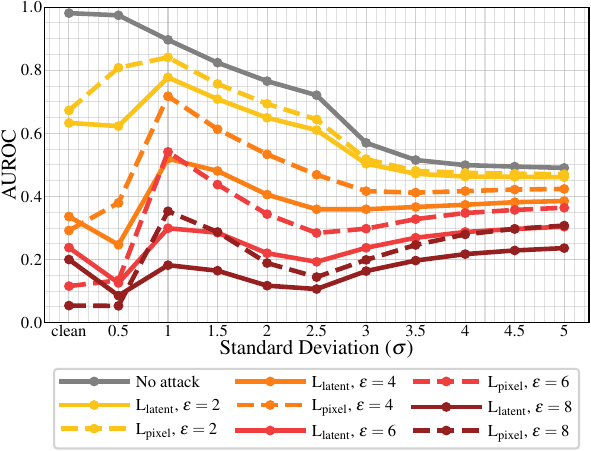}
\caption{Blur}
\end{subfigure}
\caption{The mean performance of all three detectors measured in AUROC under the proposed attack with various $\epsilon$-budgets under JPEG and Blur.}
\label{fig:defenses}
\end{figure}

The mean performance of the three selected detectors under the proposed attacks with $\epsilon=2, 4, 6, 8$ in the presence of JPEG compression and Gaussian blur is summarized in \fig{defenses}. The attack using \lossA is always represented by a solid line, whereas \lossB is represented by a dashed line.

As expected, applying distortions decreases the baseline performance of the detectors. Interestingly, the performance under all the attacks slightly improves when the final image is distorted only by a small amount. In particular, applying JPEG with $q=90$ gives a small boost in performance due to the perturbations being distorted by the compression, reducing their impact. However, this protective effect is eventually outweighed as the compression level increases. A similar effect is observed with blur.
In general, a distorted adversarially perturbed fake image is still misclassified when compared to a distorted real image.

\subsection{Ablations}

The following ablations investigate the influence of the distance function used for the \lossB attack and the $S_\text{latent}$ component of RDD.

\subsubsection{MSE as Pixel Distance}

 \begin{figure}[t]
\centering
\includegraphics[width=0.49\linewidth]{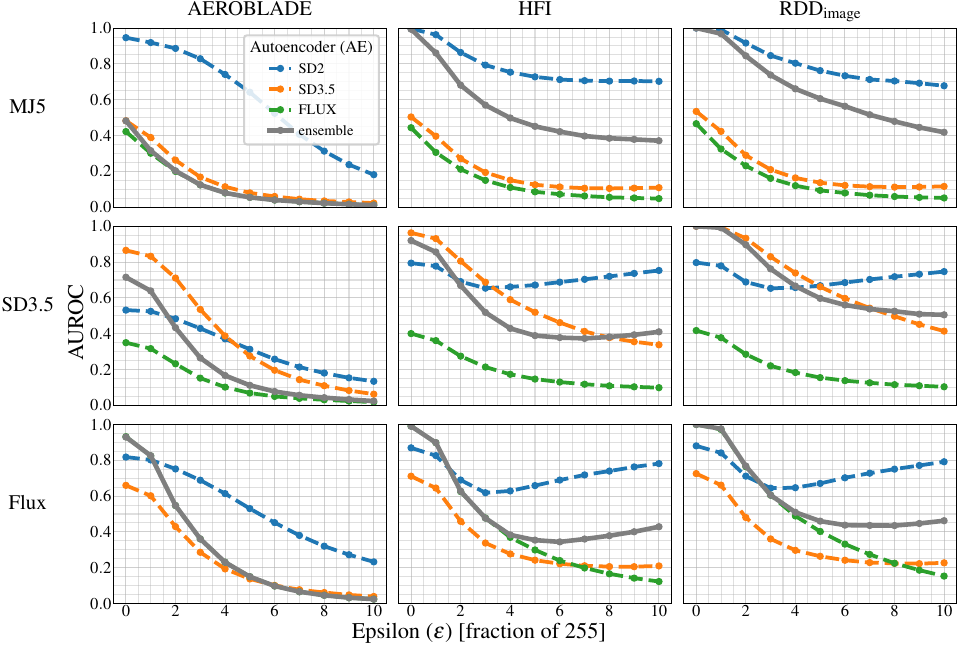}
\caption{AUROC of the detectors under the proposed attack on the three selected datasets using MSE distance.}
\label{fig:ablation1}
\end{figure}

The proposed \lossB attack uses LPIPS in its attack objective to match the distance used by the evaluated detectors. To investigate the versatility of the attack, we replace it with the mean squared error (MSE) as follows:

\begin{equation}
L_\text{MSE}(x, \eta) = \text{MSE}(x+\eta, \mathcal{D} \circ \mathcal{E} (x+\eta)) \enspace,
\end{equation}

As shown in \fig{ablation1}, the adversarial effect is considerably weaker than with LPIPS. However, the attack remains effective for AEROBLADE and for all configurations of HFI and RDD except when only SD2 is used for detection. This demonstrates that the attack is not limited by the choice of the specific distance function used by the target detector.

\subsubsection{Full RDD}

Although LDM-generated images are generally detected using the $S_\text{image}$ score of RDD, it can be of interest to observe if the $S_\text{latent}$ score can improve detection of perturbed images. The performance of RDD is evaluated on 100 images from the Midjourney dataset, using the SD2 AE and U-Net for detection and the SD2 AE as the attack target.

 \begin{figure}[t]
\centering
\begin{subfigure}{0.49\columnwidth}
\centering
\includegraphics[width=\linewidth]{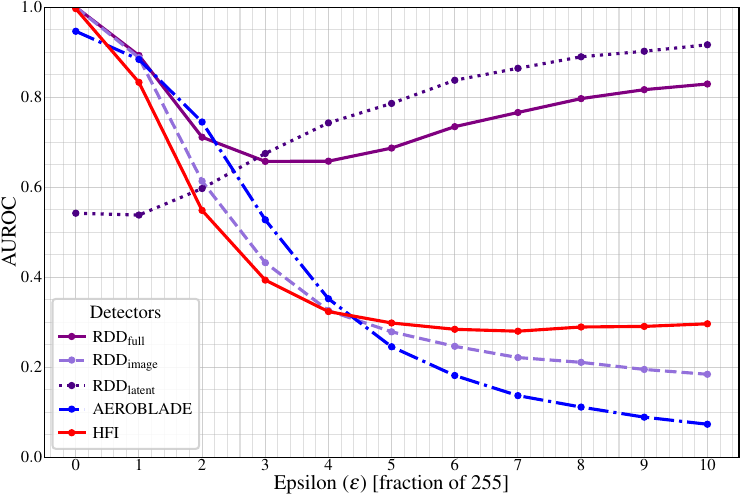}
\caption{\lossA}
\end{subfigure}
\hfill
\begin{subfigure}{0.49\columnwidth}
\centering
\includegraphics[width=\linewidth]{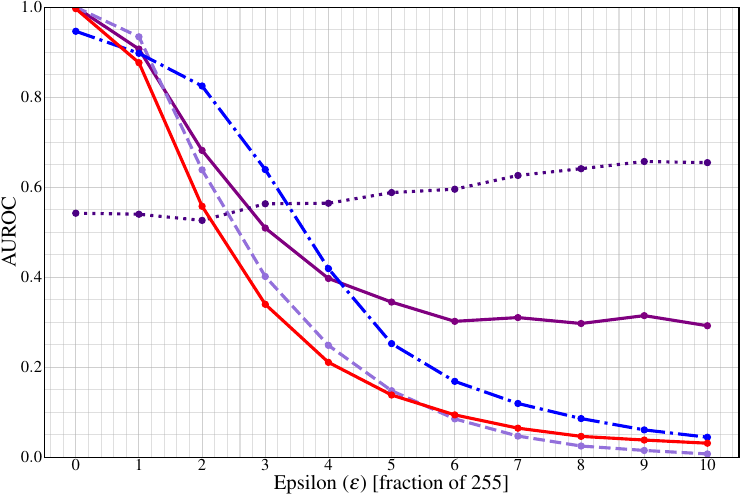}
\caption{\lossB}
\end{subfigure}
\caption{AUROC of the detectors using the SD2 AE under the proposed attack on the Midjourney dataset.}
\label{fig:ablation2}
\end{figure}

As shown in \fig{ablation2}, the $S_\text{latent}$ score actually improves as perturbation strength increases. At low perturbation strengths, its AUROC is close to 0.5, as both real and Midjourney images can be reconstructed easily by the SD2 diffusion model. Then, as the perturbations move the images away from the learned manifold of SD2, the reconstruction error increases, and $S_\text{latent}$ becomes a more reliable detection method for such perturbed images. The contribution of $S_\text{latent}$ improves the overall RDD score, reducing the strength of the attack. Interestingly, this effect is significant for \lossA but is considerably weaker for \lossB.

\section{Discussion and Limitations}

Our work reveals an inherent vulnerability of training-free reconstruction-based detectors.
Across different datasets, AEs, and detectors, we observe that adversarial perturbations significantly decrease detection performance.
Moreover, our results confirm our assumptions described in Section~\ref{sec:threat}.
Regardless of whether the AE used to craft the adversarial example is part of the defender's ensemble, the attack remains effective.
In addition, attacks are not limited to a specific detector, but apply to \emph{all} detectors leveraging AE-based reconstruction errors.
Together, this means that attacks will likely be effective against future reconstruction-based detectors, as long as they employ the same core methodology.

Despite carefully choosing our experimental conditions, our work has several limitations. 
First and foremost, our attacks are directly targeted at reconstruction-based detectors and may not generalize to classifier-based detectors.
Moreover, although we observe transferability, some settings we consider assume that the attacker has access to the AE that generated their image.
While this is naturally given for open-source models, attacks using proprietary generative models may be less effective.
Due to the computational complexity and the large number of possible combinations between datasets, AEs, and detectors, the number of images we analyzed per experiment is relatively small.

While our work uncovers a fundamental vulnerability of reconstruction-based detectors, there are several aspects left for future work.
The most pressing one is the development of effective defenses.
Another interesting research direction may be creating adversarial examples that are effective against both classifier- and reconstruction-based detectors at the same time.

\section{Conclusion}

In contrast to conventional data-driven detection approaches, reconstruction-based detectors have several desirable properties.
They do not require any training, can be easily extended to new generative models, and offer interpretable detection and attribution.
These properties are a consequence of leveraging the model's own AE for calculating reconstruction errors.
In this work, we demonstrate a fundamental downside of this approach: By adding imperceptible adversarial perturbations, we can intentionally increase the difference between original and reconstruction, rendering the detector ineffective.
Moreover, the fact that the AEs leveraged for detection are usually openly available benefits the attacker, mimicking a natural white-box setting.
We call upon the research community to critically revisit the idea of reconstruction-based AI-generated image detection and consider its adversarial robustness as an important direction for future work.

\bibliographystyle{splncs04}
\bibliography{main}
\end{document}

%% file: preamble.tex
\newcommand{\aeix}[2]{\text{AE}_{#1}(#2)}

\newcommand{\adv}[1]{#1'}
\newcommand{\enc}[0]{\mathcal{E}}

\newcommand{\enci}[2]{\enc_{#1}(#2)}

\newcommand{\lossA}[0]{$L_{\text{latent}}$\xspace}
\newcommand{\lossB}[0]{$L_{\text{pixel}}$\xspace}
\newcommand{\fig}[1]{Figure~\ref{fig:#1}}